\documentclass{bmvc2k}

\usepackage{booktabs}       %
\usepackage{multirow}
\usepackage{amsmath,amssymb}
\usepackage{color}
\usepackage{caption}
\usepackage{graphicx}
\usepackage{xspace}
\usepackage{bbm}
\usepackage{makecell}
\usepackage{nccmath}

\newcommand{\mc}{\mathcal}
\newcommand{\R}{\mathbb{R}}

\newcommand{\D}{\mathcal{D}}
\newcommand{\x}{\boldsymbol{x}}

\newcommand{\norm}[1]{\left|\left|#1\right|\right|}
\newcommand{\softmax}{\textnormal{softmax}}

\makeatletter
\DeclareRobustCommand\onedot{\futurelet\@let@token\@onedot}

\def\@onedot{\ifx\@let@token.\else.\null\fi\xspace}

\def\ie{\emph{i.e}\onedot}

\def\etal{\emph{et al}\onedot}

\title{Surprisingly Simple Semi-Supervised Domain Adaptation with Pretraining and Consistency}

\addauthor{Samarth Mishra}{samarthm@bu.edu}{1}
\addauthor{Kate Saenko}{saenko@bu.edu}{12}
\addauthor{Venkatesh Saligrama}{srv@bu.edu}{1}

\addinstitution{
 Boston University
}
\addinstitution{
 MIT-IBM Watson AI Lab
}

\runninghead{Mishra, Saenko, Saligrama}{SSDA with Pretraining and Consistency}

\def\etal{\emph{et al}\bmvaOneDot}

\begin{document}

\maketitle
\begin{abstract}
Most modern unsupervised domain adaptation (UDA) approaches are rooted in domain alignment, \ie, learning to align source and target features to learn a target domain classifier using source labels. In semi-supervised domain adaptation (SSDA), when the learner can access few target domain labels, prior approaches have followed UDA theory to use domain alignment for learning. We show that the case of SSDA is different and a good target classifier can be learned without needing alignment. We use self-supervised pretraining (via rotation prediction) and consistency regularization to achieve well separated target clusters, aiding in learning a low error target classifier. With our Pretraining and Consistency (PAC) approach, we achieve state of the art target accuracy on this semi-supervised domain adaptation task, surpassing multiple adversarial domain alignment methods, across multiple datasets. PAC, while using simple techniques, performs remarkably well on large and challenging SSDA benchmarks like DomainNet and Visda-17, often outperforming recent state of the art by sizeable margins. 
Code for our experiments can be found at \href{https://github.com/venkatesh-saligrama/PAC}{https://github.com/venkatesh-saligrama/PAC}.

\end{abstract}
\section{Introduction}

The problem of visual domain adaptation arises when a learner must leverage labeled source domain data to classify instances in the target domain, where it has limited access to ground-truth annotated labels. An example of this is the problem of learning to classify real-world images based on hand-sketched depictions. The problem is challenging because discriminative features that are learnt while training to classify source domain instances may not be meaningful or sufficiently discriminative in the target domain. As described in prior works, this situation can be viewed as arising from a ``domain-shift'', where the joint distribution of features and labels in the source domain does not follow the same law in the target domain. 

\begin{figure}[t!]
    \centering
    \includegraphics[width=0.75\linewidth]{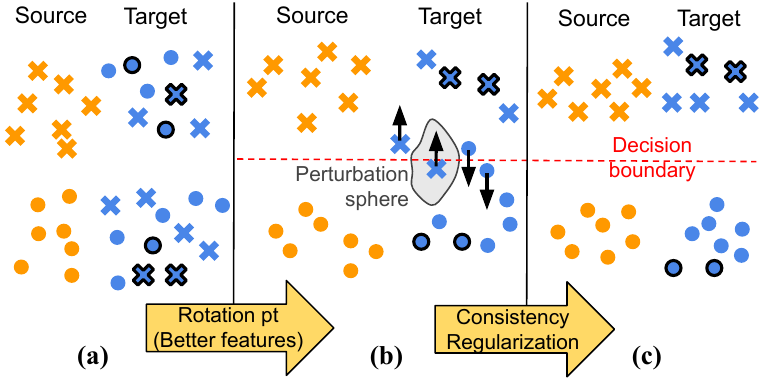}
    \vspace{1mm}
    \caption{\label{fig:hard} Feature space illustration of our approach. Shapes of points represent classes and colors represent different domains. Target domain points with a black border are labelled. Initial feature space might not be discriminative on target causing subsequent errors with domain alignment. Pretraining with rotation prediction can help fix initial features to some extent and cluster them better by class (a $\rightarrow$ b).  Consistency regularization helps cluster features by making the classifier invariant to input perturbations, keeping decision boundaries away from data-dense regions. Image space perturbations can help correctly cluster points that intially lie on the wrong side of the boundary (b $\rightarrow$ c).}
    \vspace{-5mm}
\end{figure}

We propose a novel method for semi-supervised domain adaptation (SSDA), where the learner, in addition to unlabeled target examples, is granted access to a few labeled target domain examples for training. Our method is based on enhancing clusterability of target domain features independent of the source domain. Prior SSDA methods \cite{saito2019semi, jiangbidirectional, kim2020attract} draw upon approaches developed within the context of visual domain adaptation using labelled source images and unlabelled target images \cite{ajakan2014domain, long2018conditional, saito2018maximum, tzeng2017adversarial, zhang2019bridging}. These works in turn are based on adversarial domain alignment, and draw upon Ben-David~\etal~\cite{ben2010theory}'s elegant theory of domain adaptation. Generalization bounds in Ben-David~\etal~\cite{ben2010theory} suggest that, good adaptation is possible if the divergence between induced source and target domain feature distributions is small. 

While domain alignment is a meaningful goal in the absence of labels, we believe that the situation is dramatically different in the presence of few target domain labels. For our work, we draw inspiration from Castelli and Cover \cite{castelli1996relative}, who show that, in binary classification problems, a learner with no knowledge of underlying feature distributions, can benefit exponentially from labelled examples, and the learner's error probability approaches Bayes error. As such, our key insight is that we can forego domain alignment if we learn representations that result in compact clusters in the target domain. The identity of these clusters can then be deduced from few-shot labels. 

Our proposed approach, PAC (pretraining and consistency), is based on combining well-known methods and novel objectives, which together lead to enhanced clusterability. In particular, we use rotation prediction pre-training \cite{gidaris2018unsupervised}, label consistency \cite{sajjadi2016regularization}, and cross-entropy losses on labelled source and target data, which together result in more compact and pure target domain clusters. Cross-entropy loss on target-domain labelled examples implies good feature separation for different target classes. Consistency loss on perturbed examples and pretraining result in improving similarity between labelled target examples and unlabelled examples. Together these objectives are effective in realizing compact clusters, and result in improved prediction. Notably, PAC achieves better target accuracy than most comparable state-of-the-art methods on the large and challenging DomainNet \cite{peng2019moment} benchmark by 3-5 \%, and on VisDA-17, it is better by 4-10\%. %

\noindent {\it Pretraining.} Following recent domain adaptation methods we also utilize an Imagenet \cite{imagenet_cvpr09} pretrained backbone as a starting point. We found that these features are somewhat biased towards Imagenet images and to enhance them, we use self-supervision via rotation prediction. Our approach is informed by Gidaris~\etal~\cite{gidaris2018unsupervised}'s work for learning semantically meaningful features from unlabelled images. In addition, it was recently found in a study by Wallace~\etal~\cite{wallace2020extending} to be more semantically meaningful than an array of other self-supervision objectives. Most importantly for us, we found that self-supervision via rotation prediction also resulted in better target clusterability as compared to Imagenet pretraining. Fig \ref{fig:hard} (a), shows a hypothetical scenario where simply aligning features can lead to poor generalization, and how pretraining can somewhat remedy this situation.

\noindent {\it Label Consistency.} Also key to our approach is label consistency using image space perturbations. Label consistency or \emph{consistency regularization} enforces invariance of classifier predictions to small input perturbations, and thus to inputs from neighborhoods that may form clusters\footnote{Fig \ref{fig:hard} depicts consistency regularization helping correctly cluster points that may initially lie on the wrong side of the decision boundary, and that may result in errors with a domain alignment strategy.}. In our approach, we do this using image augmentation methods like RandAugment \cite{cubuk2020randaugment} and color jittering, and the model is trained to produce the same output for both a perturbed and an unperturbed version of the image resulting in meaningful neighborhood for images\footnote{label consistency is related to cluster assumption \cite{chapelle2005semi}, which suggests similar labels for examples within a cluster, as well as the concept of \emph{low-density separation assumption}\cite{verma2019interpolation}, which requires classifier decision boundaries to lie in low-density regions of feature space.}. %

Our contributions in this paper are two-fold:
\begin{itemize}
    \item We propose a novel semi-supervised domain adaptation method PAC, based on label consistency and rotation prediction for pretraining, which performs comparably or better than state of the art on SSDA across multiple datasets. In contrast to prior works, we forego domain alignment, and pose objectives that improve target clusterability.
    \item We perform ablative analysis on individual components of our method, illustrating their behavior, and their impact on performance. Our analysis provides an understanding of these components and shows how they can be combined with other techniques.
\end{itemize}

\section{Background} \label{sec:background}

\noindent\textbf{Adversarial Domain Alignment.}
The generalization bound for target error for unsupervised domain adaptation scenario in Ben-David~\etal~\cite{ben2010theory} is a sum of three terms: the empirical source error; the divergence between source and target distributions $X_S$ and $X_T$ respectively; and the smallest error of a jointly trained classifier. In the absence of labels, this bound is somewhat intuitive, namely that a joint classifier is required, and that such a classifier in general would suffer from distributional differences.  
Motivated by this theory, a range of visual domain adaptation approaches attempt to find a feature space $\phi$ such that the distributions $\phi(X_S)$ and $\phi(X_T)$ have low divergence, so that the distributional differences are mitigated. \cite{ben2010theory}'s distributional difference is measured in terms of the worst-case differences in disagreement of any two classifiers on source and target spaces. As such the goal of minimizing distributional alignment results in a minimax objective, which, along with the minimization of a classifier error on source labels is broadly the approach adopted by a range of recent domain adaptation methods \cite{ajakan2014domain, ganin2015unsupervised, long2018conditional, saito2018maximum, tzeng2017adversarial, zhang2019bridging}.

\noindent\textbf{The Exponential Value of Labels.} 
Castelli \& Cover \cite{castelli1995exponential, castelli1996relative} showed that for a conventional binary classification problem with $u$ unlabelled examples and $\ell$ labelled examples, the learner's probability of error approaches the Bayes error as 
{\small
\begin{gather} \label{eq.labelvalue}
P_{err} - P_{bayes} = O \left(\frac{1}{u} \right) + \exp\left(-D(q_1,q_2)\ell + o(\ell)\right)   
\end{gather}}
\noindent where $q_1$ and $q_2$ are feature distributions conditioned on class-labels, and unknown to the learner. $D(q_1,q_2)$ %
is the Bhattacharyya Distance between the distributions. A key insight is that the error probability decays exponentially with the product of inter-class distance and number of labels. Thus low error is achievable with only a few labels for a feature distribution that is well separated across classes. This is the key motivation behind our approach.

For learning a good target classifier, we formulate an objective that, in addition to penalizing source and target domain mistakes on labelled data, improves separability of features belonging to different target classes. %
Informally, our objective is to optimize\\
{\it Source Loss + Target Loss + Target Clustering.}\\
The first term leverages source data to realize good feature representation. On the other hand, the second and third terms, together attempt to improve target domain representation. As such target loss promotes good feature separation for labelled examples between different target classes. We utilize label consistency to improve target clustering by enforcing target invariance of classifier predictions to small input perturbations, and thus to inputs from neighborhoods that may form clusters. Target clustering is also aided by our initialization with self-supervised pretraining. Together, the two terms result in operationalizing \cite{castelli1995exponential}'s concept resulting in well-separated clusters, and allow for generalization from few labels.

\section{Pretraining and Consistency (PAC)}

\begin{figure*}[t!]
    \centering
    \includegraphics[width=0.85\linewidth,trim=0cm 0cm 0cm 0cm,clip]{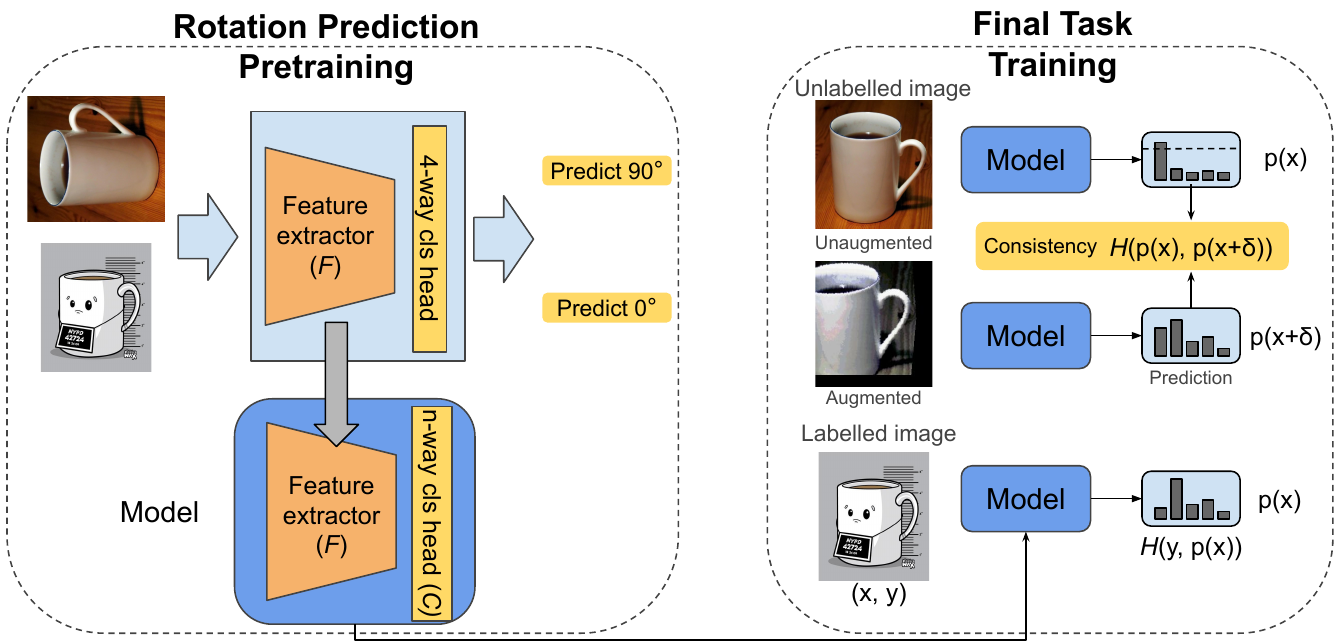}
    \vspace{3mm}
    \caption{A diagram of our Pretraining and Consistency (PAC) approach. We first train our backbone for the self-supervised task of predicting rotations (left). This backbone is then used as a warm start for our classification model, which uses labelled data with the cross entropy criterion and consistency regularization for the unlabelled data (right).}
    \vspace{-5mm}
    \label{fig:model}
\end{figure*}

Before describing our approach, we introduce some notation for ease of precise exposition. Available to the model are two sets of labelled images : $\D_s = \{(\x_i^s, y_i^s)\}_{i=1}^{n_s}$, the labelled source images and $\D_t = \{(\x_i^t, y_i^t)\}_{i=1}^{n_t}$, the few labelled target images, and additionally the set of unlabelled target images $\D_u = \{x_i^u\}_{i=1}^{n_u}$. The goal is to predict labels for these images in $\D_u$. The final classification model consists of two components : the feature extractor $F$ and the classifier $C$. $F$ generates features $F(\x)$ for an input image $\x$ which the classifier operates on to produce output class scores $C(F(\x)) \in \R^{K}$, where $K$ is the number of categories that the images in the dataset could belong to. In our experiments, $F$ is a convolutional network and produces features with unit $\ell_2$-norm, \ie $\norm{F(x)}_2 = 1$ (following \cite{saito2019semi}). $C$ consists of one or two fully connected layers.

\noindent An overview of PAC is shown in Fig \ref{fig:model}. Our final model is trained in two stages:

\noindent \textbf{Pretraining with Rotation Prediction.} We first train our feature extractor $F$ with the self-supervised task of predicting image rotations (Fig \ref{fig:model} (left)) on both the source and target datasets, \ie, all images in $\D_s, \D_t$ and $\D_u$. Without using image category labels, we train a 4-way classifier to predict one out of 4 possible angles ($0^\circ$, $90^\circ$, $180^\circ$, $270^\circ$) that an input image has been rotated by. We follow the procedure of Gidaris~\etal~\cite{gidaris2018unsupervised} and in each minibatch, we introduce all 4 rotations of a given image to the classifier. This backbone is then used as the initialization for training the final classifier in the next stage.

\noindent \textbf{Consistency Regularized Classifier Training.} 
Consistency regularization promotes the final model $C \circ F$ to produce the same output for both an input image $\x$ and a perturbed version $\x + \delta$. We introduce these perturbations using image level augmentations --- RandAugment \cite{cubuk2020randaugment} along with additional color jittering. Given an unlabelled image $\x \in \D_u$, we first compute the model's predicted class distributions 
\begin{gather*}
    p_x = p(y | \x; F, C) = \softmax(C(F(\x))) \\
    q_x = p(y | \x + \delta; F, C) = \softmax(C(F(\x + \delta)))
\end{gather*}
$p_x$ is then confidence thresholded using a threshold $\tau$ and the following is used as the consistency regularization loss.
\begin{align}
    \mc L_{CR}(\x) = \mathbbm{1}[\max_{k \in K} p_x(k) \ge \tau] H(p_x, q_x)
\end{align}
where $\mathbbm{1}$ is an indicator function and $H(p_x, q_x) = \sum_{k \in K} -p_x(k)\log (q_x(k))$ is cross-entropy. Note that $p_x(\cdot)$ has been used to index into the $K$ elements of $p_x$. Intuitively, an unperturbed version of image $\x$ is used to compute \emph{pseudo-targets} for the perturbed version $\x + \delta$, which is only used when the pseudo-target has high confidence ($\max_{k} p_x(k) \ge \tau$). We also note here that the target $p_x$ is treated as a constant for gradient computation with respect to the network parameters. For the labelled examples from $\D_s$ and $\D_t$, we use the same perturbations but with ground truth labels as targets.

The model is optimized using minibatch-SGD, with minibatches $M_s, M_t$ and $M_u$ sampled from $\D_s, \D_t$ and $\D_u$ respectively. The final optimization criterion used is
\begin{align*}
    \mc L = \frac{1}{|M_s|} \sum_{(\x, y) \in M_s} & H(\bar{y}, \x) + \frac{1}{|M_t|} \sum_{(\x, y) \in M_t} H(\bar{y}, \x) \\ 
    &+ \frac{1}{|M_u|} \sum_{\x \in M_u} \mc L_{CR} (\x)
\end{align*}
where $\bar{y} \in \R^{K}$ is the one-hot representation of $y \in [K]$ or $\bar{y}(i) = \mathbbm{1} [i=y]$ and $H(\bar{y}, \x)$ has been overloaded to mean $H(\bar{y}, C(F(\x)))$.

\section{Experiments} \label{sec:expts}

\noindent \textbf{Datasets.}
We evaluated PAC on four different datasets: DomainNet \cite{peng2019moment}, VisDA-17 \cite{peng2017visda}, Office-Home \cite{venkateswara2017Deep} and Office \cite{saenko2010adapting}. DomainNet \cite{peng2019moment} is a recent large scale domain adaptation benchmark with 6 different visual domains and 345 classes. We use a subset of 4 domains (Clipart, Paintings, Real, and Sketch) and 126 classes for our experiments. This subset had close to 36500 images per domain. A total of 7 different scenarios out of the possible 12 were used for evaluation. VisDA-17 is another large benchmark with a single adaptation scenario : the source domain (the VisDA "train" split) consists of 152,398 synthetic images from 12 categories, and the target domain (the VisDA "validation" split) consists of 55,388 real images.  Office-Home \cite{venkateswara2017Deep} and Office \cite{saenko2010adapting} datasets are relatively smaller in size with images of objects typically found in office and home environments. More details and the performance comparison for these datasets are in Appendix \ref{sec:full-results}.

For each adaptation scenario, following \cite{saito2019semi}, we use 1-shot and 3-shot settings for evaluation, where 1 and 3 target labels per class are available to the learner respectively (See appendix \ref{sec:diff-shots} for an analysis varying the number of target shots). For each scenario, 3 examples in the target domain are held out for validation, except in VisDA-17 where 20 examples per class were held out because of the fewer number of categories.

\noindent \textbf{Implementation Details.} All our experiments were implemented in PyTorch \cite{pytorch} using W\&B \cite{wandb} for tracking experiments. For evaluations on the DomainNet dataset, we used an Alexnet and a Resnet-34 \cite{he2016deep} backbone, while on VisDA-17 we evaluated our method with a ResNet-34 backbone. 1 or 2 fully connected layers we used for the classifier $C$. Optimization was done via SGD with a momentum 0.9. Complete details of all experiments are in Appendix \ref{sec:expt-deets}.

\subsection{Results} \label{subsec:results}

\begin{table*}[t]
\begin{center}
\scalebox{0.6}{
\begin{tabular}{l|l|cccccccccccccc|cc}
\toprule[1.5pt] %
 \multirow{2}{*}{Net} & \multirow{2}{*}{Method}       &\multicolumn{2}{c}{R to C}&\multicolumn{2}{c}{R to P} & \multicolumn{2}{c}{P to C}  & \multicolumn{2}{c}{C to S} & \multicolumn{2}{c}{S to P} & \multicolumn{2}{c}{R to S} & \multicolumn{2}{c}{P to R}     &\multicolumn{2}{|c}{MEAN} \\ %
& &1\scriptsize{-shot}&3\scriptsize{-shot} &1\scriptsize{-shot}&3\scriptsize{-shot}&1\scriptsize{-shot}&3\scriptsize{-shot} &1\scriptsize{-shot}&3\scriptsize{-shot}&1\scriptsize{-shot}&3\scriptsize{-shot}&1\scriptsize{-shot}&3\scriptsize{-shot} &1\scriptsize{-shot}&3\scriptsize{-shot} &1\scriptsize{-shot}&3\scriptsize{-shot}  \\ \hline
 \multirow{9}{*}{Alexnet} 
 & S+T       & 43.3 & 47.1 & 42.4 & 45.0 & 40.1 & 44.9 & 33.6 & 36.4 & 35.7 & 38.4 & 29.1 & 33.3 & 55.8 & 58.7 & 40.0 & 43.4 \\  
 & DANN      & 43.3 & 46.1 & 41.6 & 43.8 & 39.1 & 41.0 & 35.9 & 36.5 & 36.9 & 38.9 & 32.5 & 33.4 & 53.6 & 57.3 & 40.4 & 42.4 \\
 & ADR      & 43.1 & 46.2 & 41.4 & 44.4 & 39.3 & 43.6 & 32.8 & 36.4 & 33.1 & 38.9 & 29.1 & 32.4 & 55.9 & 57.3 & 39.2 & 42.7 \\
 & CDAN     & 46.3 & 46.8 & 45.7 & 45.0 & 38.3 & 42.3 & 27.5 & 29.5 & 30.2 & 33.7 & 28.8 & 31.3 & 56.7 & 58.7 & 39.1 & 41.0 \\
 & MME      & 48.9 & 55.6 & 48.0 & 49.0 & 46.7 & 51.7 & 36.3 & 39.4 & 39.4 & 43.0 & 33.3 & 37.9 & 56.8 & 60.7 & 44.2 & 48.2 \\
 & Meta-MME & -    & 56.4 & -    & 50.2 & -    & 51.9 & -    & 39.6 & -    & 43.7 & -    & 38.7 & -    & 60.7 & -    & 48.7 \\
 & APE      & 47.7 & 54.6 & 49.0 & 50.5 & 46.9 & 52.1 & 38.5 & 42.6 & 38.5 & 42.2 & 33.8 & 38.7 & \textbf{57.5} & \textbf{61.4} & 44.6 & 48.9 \\
 & BiAT     & 54.2 & 58.6 & 49.2 & 50.6 & 44.0 & 52.0 & 37.7 & 41.9 & \textbf{39.6} & 42.1 & 37.2 & 42.0 & 56.9 & 58.8 & 45.5 & 49.4 \\
 & CDAC & \textbf{56.9} & \textbf{61.4} & \textbf{55.9} & \textbf{57.5} & \textbf{51.6} & \textbf{58.9} & \textbf{44.8} & \textbf{50.7} & \textbf{48.1} & \textbf{51.7} & \textbf{44.1} & \textbf{46.7} & \textbf{63.8} & \textbf{66.8} & \textbf{52.1} & \textbf{56.2} \\
 & PAC     & \textbf{55.4} & \textbf{61.7} & \textbf{54.6} & \textbf{56.9} & \textbf{47.0} & \textbf{59.8} & \textbf{46.9} & \textbf{52.9} & 38.6 & \textbf{43.9} & \textbf{38.7} & \textbf{48.2} & 56.7 & 59.7 & \textbf{48.3} & \textbf{54.7} \\ \midrule \midrule

\multirow{9}{*}{Resnet-34}     
 & S+T  & 55.6 & 60.0 & 60.6 & 62.2 & 56.8 & 59.4 & 50.8 & 55.0 & 56.0 & 59.5 & 46.3 & 50.1 & 71.8 & 73.9 & 56.8 & 60.0 \\
 & DANN      & 58.2 & 59.8 & 61.4 & 62.8 & 56.3 & 59.6 & 52.8 & 55.4 & 57.4 & 59.9 & 52.2 & 54.9 & 70.3 & 72.2 & 58.4 & 60.7 \\
 & ADR      & 57.1 & 60.7 & 61.3 & 61.9 & 57.0 & 60.7 & 51.0 & 54.4 & 56.0 & 59.9 & 49.0 & 51.1 & 72.0 & 74.2 & 57.6 & 60.4 \\
 & CDAN     & 65.0 & 69.0 & 64.9 & 67.3 & 63.7 & 68.4 & 53.1 & 57.8 & 63.4 & 65.3 & 54.5 & 59.0 & 73.2 & 78.5 & 62.5 & 66.5 \\
 & MME      & 70.0 & 72.2 & 67.7 & 69.7 & 69.0 & 71.7 & 56.3 & 61.8 & 64.8 & 66.8 & 61.0 & 61.9 & 76.1 & 78.5 & 66.4 & 68.9 \\
 & Meta-MME & -    & 73.5 & -    & 70.3 & -    & 72.8 & -    & 62.8 & -    & 68.0 & -    & 63.8 & -    & 79.2 & -    & 70.1 \\
 & APE      & 70.4 & 76.6 & 70.8 & 72.1 & \textbf{72.9} & \textbf{76.7} & 56.7 & 63.1 & 64.5 & 66.1 & 63.0 & 67.8 & 76.6 & \textbf{79.4} & 67.8 & 71.7 \\
 & BiAT     & 73.0 & 74.9 & 68.0 & 68.8 & 71.6 & 74.6 & 57.9 & 61.5 & 63.9 & 67.5 & 58.5 & 62.1 & \textbf{77.0} & 78.6 & 67.1 & 69.7 \\
 & CDAC & \textbf{77.4} & \textbf{79.6} & \textbf{74.2} & \textbf{75.1} & \textbf{75.5} & \textbf{79.3} & \textbf{67.6} & \textbf{69.9} & \textbf{71.0} & \textbf{73.4} & \textbf{69.2} & \textbf{72.5} & \textbf{80.4} & \textbf{81.9} & \textbf{73.6} & \textbf{76.0} \\
 & PAC     & \textbf{74.9} & \textbf{78.6} & \textbf{73.0} & \textbf{74.3} & 72.6 & 76.0 & \textbf{65.8} & \textbf{69.6} & \textbf{67.9} & \textbf{69.4} & \textbf{68.7} & \textbf{70.2} & 76.7 & 79.3 & \textbf{71.4} & \textbf{73.9} \\ \hline
\bottomrule[1.5pt]
\end{tabular}}
\end{center}
\caption{Accuracy on the DomainNet dataset (\%) for one-shot and three-shot settings on 4 domains, R: Real, C: Clipart, P: Painting, S: Sketch. PAC, though simple, is strong enough to be competitive with or outperform other state of the art approaches on most scenarios. Top 2 accuracies in each column are highlighted in bold}
\vspace{-3mm}
\label{tab:domainnet}
\end{table*}

\begin{table}[h]
\begin{center}
\begin{tabular}{lcccc}
\toprule[1.5pt]
\multirow{2}{*}{Method} & \multicolumn{4}{c}{Overall Accuracy} \\ 
& 1-shot & 3-shot & 1-pct & 5-pct \\\midrule
S+T & 57.7 & 59.9 & 76.2 & 82.9 \\
MME & 69.7 & 70.7 & 80.5 & 84.1 \\
LIRR & - & - & 81.7 & 84.5 \\
LIRR+CosC & - & - & 82.3 & 85.1 \\
PAC & \textbf{75.2} & \textbf{80.4} & \textbf{86.0} & \textbf{88.9} \\
\bottomrule[1.5pt]
\end{tabular}
\end{center}
\caption{Results on VisDA-17. PAC outperforms MME and LIRR, both current state of the art approaches, by a sizeable margin. CosC stands for cosine classifier and ``1-pct'' and ``5-pct'' denote scenarios where 1\% and 5\% of the target images are labelled respectively. \label{tab:visda17}}
\vspace{-5mm}
\end{table}

\noindent \textbf{Comparison to other approaches.} We compare PAC with different recent SSDA approaches : MME \cite{saito2019semi}, BiAT \cite{jiangbidirectional}, Meta-MME \cite{li2020online}, APE \cite{kim2020attract}, LIRR \cite{li2021learning}, and CDAC \cite{li2021cross} using results reported by these papers. Besides this, we also include in the tables, baseline approaches using adversarial domain alignment---DANN \cite{ganin2016domain}, ADR \cite{saito2017adversarial} and CDAN \cite{long2018conditional}, that were evaluated by Saito~\etal~\cite{saito2019semi}. The baseline ``S+T'' is a method that simply uses all labelled data available to it to train the network using cross-entropy loss. Note that PAC can be construed as ``S+T'' along with additional consistency regularization and with a warm start using rotation prediction for pretraining.

In Table \ref{tab:domainnet}, we compare the accuracy of PAC with different recent approaches on DomainNet. Remarkably our simple approach outperforms other SSDA approaches by 3-5\% on this benchmark with different backbones (with the exception of CDAC, with which it is competitive on most scenarios). In Table \ref{tab:visda17} holding the VisDA-17 results, besides our method, we report results of S+T and MME, that we replicated from the implementation of \cite{saito2019semi}, on the 1-shot and 3-shot scenarios. In addition, for comparison with LIRR on equal footing, we evaluated PAC on scenarios with 1\% and 5\% of all the target examples are labelled. We see that PAC shows strong performance, with close to 10\% improvement in accuracy over MME in the 3-shot scenario, and 4\% improvement over LIRR.

\noindent \textbf{Ablative analysis.} In Table \ref{tab:ablation}, we see what rotation prediction pretraining and consistency regularization do for final target classification performance separately. The two components provide boosts to the final performance individually, with the combination of both performing best. We see that in most cases consistency regularization helps performance significantly, especially in the 3-shot scenarios.

\begin{table}[h]
\begin{center}
\begin{tabular}{cccccc}
\toprule[1.5pt] %
 \multirow{3}{*}{Rot$^n$} & \multirow{3}{*}{CR}  & \multicolumn{4}{c}{Target Accuracy} \\
 & & \multicolumn{2}{c}{Alexnet} & \multicolumn{2}{c}{Resnet-34} \\
 & & 1\scriptsize{-shot} & 3\scriptsize{-shot} & 1\scriptsize{-shot} & 3\scriptsize{-shot} \\ \midrule
 &  &                       29.1 & 33.3 & 46.3 & 50.1 \\
 \checkmark &  &             35.1 & 37.9 & 54.1 & 56.1 \\
 & \checkmark &              32.5 & 45.9 & 64.3 & 68.9 \\
 \checkmark & \checkmark &   38.7 & 48.2 & 68.7 & 70.2 \\
\bottomrule[1.5pt]
\end{tabular}
\end{center}
\caption{Ablation study for pretraining predicting rotations (Rot$^{n}$) and consistency regularization (CR) on the \emph{real} to \emph{sketch} scenario of Domainnet using both Alexnet and Resnet-34 backbones.}
\vspace{-5mm}
\label{tab:ablation}
\end{table}

\begin{figure*}[t!]
    \centering
    \includegraphics[width=0.95\linewidth,trim=0cm 0cm 0cm 0cm,clip]{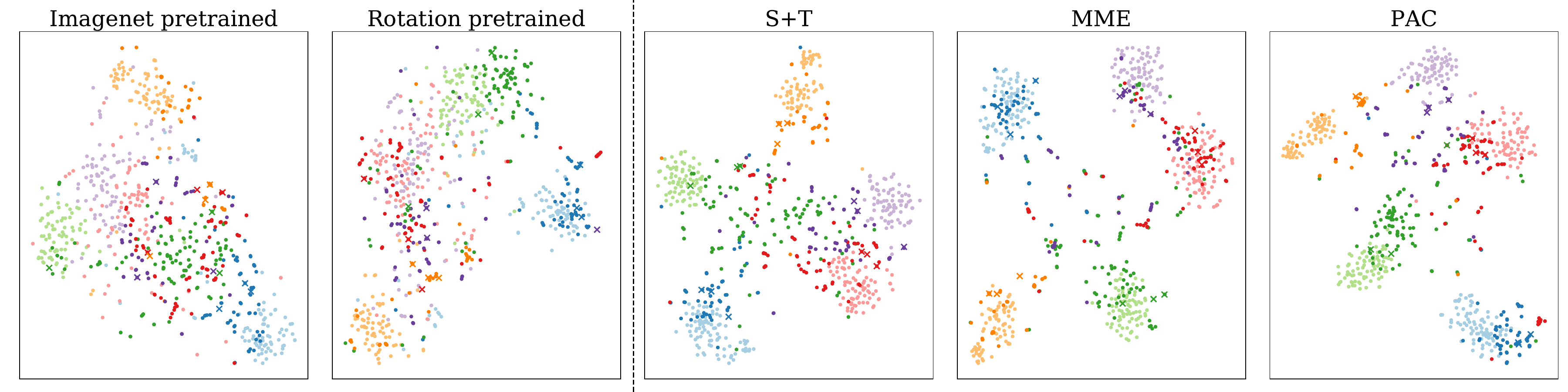}
    \vspace{1mm}
    \caption{2-D TSNE embeddings of features from 5 randomly chosen classes. Lighter colors represent source domain points and darker ones represent corresponding target domain points. The dark colored X's are labelled target domain points (3 per class). Pretraining with rotation prediction begins to cluster points of the same class a little better over a backbone pretrained only on Imagenet. Our method on the right, separates classes just as well as MME. Refer to Section \ref{subsec:results} for discussion. (Best viewed with color and under zoom)}
    \vspace{-3mm}
    \label{fig:tsne}
\end{figure*}

\noindent\textbf{Feature space analysis.} In Fig \ref{fig:tsne} we plot the 2-D TSNE \cite{maaten2008visualizing} embeddings for features generated by 5 differently trained Alexnet backbones. The embeddings are plotted for all points from 5 randomly picked classes. The source domain points which are light colored circles, come from \emph{real} images of Office-Home and the target domain points which are dark colored markers come from \emph{clipart} images. The labelled target examples are marked with X's. The two plots on the left compare differently pretrained backbones and the three on the right use backbones at the end of different SSDA training processes. In the plots we can see that pretraining the backbone for rotation prediction starts to align and cluster points according to their classes a little better than what a backbone pretrained just on Imagenet can do. Out of the final classifiers on the right, we see that both PAC and MME create well separated classes in feature space allowing for the classifier to have decision boundaries in low-density regions. MME explicitly minimizes conditional entropy which may draw samples even further apart from the classifier boundaries, as compared to our method which simply tries to ensure that the classifier does not separate an example and its perturbed version.

In Table \ref{tab:feat_space}, we quantitatively analyze the features using different metrics : The $\mc A$-distance is a distance metric between the two domains in feature space computed using an SVM trained to classify domains as done in \cite{ben2006analysis}. The higher the error of the SVM domain classifier, the lower is the $\mc A$-distance. The other two metrics are accuracies of distance based nearest neighbor (NN) classifiers in feature space. The first one, ``NN Acc. (Target)'' is the accuracy of a classifier that assigns any unlabelled target example, the class label of the target labelled examples closest to it on average in the feature space. ``NN Acc. (Source)'' similarly uses only the source examples, all of which are labelled, to compute the class label for an unlabelled target example. Finally BD is an empirical inter-class Bhattacharyya distance estimate (details in appendix \ref{subsec:expt-deets-other}). Comparing the pretrained backbones, we see that rotation pretraining improves the feature space both by bringing closer the features across the two domains (as indicated by the low $\mc A$-distance) and aligning them so that features within a class are closer (indicated by the higher NN accuracies). When it comes to final feature spaces of the SSDA methods, we see that MME, being a domain alignment method, reduces $\mc A$-distance more than PAC. However, PAC is able to better maintain the class-defined neighborhood of features, as indicated by the higher accuracies. Also, PAC has higher inter-class divergence (BD), which leads to lower error according to Eq. \ref{eq.labelvalue}. This shows that in SSDA, for learning good target classification, low divergence between domains is not a necessity.

\begin{table}[t]
\begin{center}
\begin{tabular}{ccccc}
\toprule[1.5pt] %
Backbone & $\mc A$-dist  & \thead{NN Acc. \\ (Target)} & \thead{NN Acc. \\ (Source)} & BD  \\
\midrule[1.5pt]

Imagenet pt.     & 1.57 & 26.8 & 26.0 & - \\
Rotation pt.     & 1.28 & 36.2 & 34.6 & -\\
\midrule
S+T                     & 1.49 & 43.0 & 43.3 & 2.01 \\
MME                     & 1.24 & 51.2 & 51.5 & 4.03 \\
PAC                    & 1.45 & 56.4 & 56.8 & 13.43 \\
\bottomrule[1.5pt]
\end{tabular}
\end{center}
\caption{Feature space metrics for different Alexnet backbones on the \emph{real} to \emph{clipart} adaptation scenario of Office-Home. We see that rotation prediction helps improve the initial feature space. Also amongst the SSDA methods, PAC maintains a better class defined neighborhood both within and across domains, even though the two domains are not aligned as closely as in the case of MME.}
\label{tab:feat_space}
\end{table}

\noindent\textbf{How does pretraining with rotation prediction compare to constrastive methods?} Contrastive pretraining methods \cite{he2020momentum, chen2020simple, chen2021exploring} have been shown to attain remarkable performance in learning features from unlabelled images that are useful for tasks like image recognition and object detection. We evaluate how momentum contrast (MoCo) \cite{he2020momentum} and SimSiam \cite{chen2021exploring} perform for pretraining our feature extractor on both source and target images, compared with rotation prediction. Table \ref{tab:pt_comparison} compares most of the same metrics as Table \ref{tab:feat_space} with the addition of final model (training with labels and consistency regularization) performance on target classification. We see that, both MoCo and SimSiam improve the imagenet pretrained features to some extent helping the final classification performance. However, this improvement is not as high as in the case of rotation prediction pretraining.  We see that MoCo has marginally better class-defined structure across domains, but a poorer structure in the target domain indicated by the accuracies of the distance based NN classifiers. Interestingly, we see that nearest neighbors classifiers perform much poorer in the case of SimSiam pretraining as compared to other cases, while the pretraining still benefits final SSDA accuracy. This might indicate that this pretraining helps with optimization in some manner while not clustering the initial feature space as much.

\begin{table}[t]
\begin{center}
\scalebox{0.9}{
\begin{tabular}{lcccc}
\toprule[1.5pt] %
\thead{Backbone \\ pretraining} & \thead{$\mc A$-dist}  & \thead{NN Acc. \\ (Target)} & \thead{NN Acc. \\ (Source)} & \thead{Final Acc.} \\
\midrule[1.5pt]

Imagenet           & 1.57 & 26.8 & 26.0 & 54.1 \\
MoCo               & 1.31 & 31.4 & 34.8 & 56.3 \\
SimSiam            & 1.53 & 19.3 & 17.1 & 56.6 \\
Rotation           & 1.28 & 36.2 & 34.6 & 58.8 \\
\bottomrule[1.5pt]
\end{tabular}
}
\end{center}
\caption{Feature space metrics and final method performance for differently pretrained Alexnet backbones on the \emph{real} to \emph{clipart} adaptation scenario of Office-Home. MoCo and SimSiam help over Imagenet pretraining, but not as much as rotation prediction.}
\label{tab:pt_comparison}
\vspace{-1mm}
\end{table}

\section{Related Work} \label{sec:related_work}
\noindent\textbf{Cluster Assumption.} As mentioned in the introduction, related to our approach is the cluster assumption that has been defined in prior semi-supervised learning work \cite{chapelle2005semi}, suggesting that data points in a cluster in the input space should be classified similarly. Conditional entropy minimization \cite{grandvalet2005semi} and consistency regularization \cite{sajjadi2016regularization, miyato2018virtual, sohn2020fixmatch} have been used as ways of achieving this.

\noindent\textbf{Semi-supervised Domain Adaptation.} 
As mentioned above, conditional entropy minimization has been used to enforce cluster assumption.
Saito~\etal~\cite{saito2019semi} cleverly built this into a minimax optimization problem to propose an adversarial domain alignment method, called minimax entropy (MME) that also plays by the cluster assumption. They evaluated MME on SSDA, and found that it performed better than other domain alignment approaches without conditional entropy minimization. 
Their approach uses entropy maximization with the classifier in an attempt to move the boundary close to and in between target unlabelled examples. Subsequent conditional entropy minimization using the feature extractor clusters the target unlabelled examples largely according to the ``split'' created by the classifier. Note that with this approach the neighborhood already gets defined by the classifier and if errors are made here, they are harder to fix. So, while PAC can fix errors like the ones in Fig. \ref{fig:hard} (b), MME may find it harder to do so. We demonstrate this in appendix \ref{sec:more-ques} by comparing both methods trained from a randomly initialized backbone, and find that PAC performs better.

Saito~\etal~\cite{saito2019semi} also used a benchmark that has been used by subsequent approaches for evaluation of SSDA methods. Here we describe some of them. 
APE \cite{kim2020attract} uses different feature alignment objectives, within and across domains along with a perturbation consistency objective. BiAT\cite{jiangbidirectional} uses multiple adversarial perturbation strategies and consistency losses alongside MME. 
Li~\etal~\cite{li2020online} proposed an online meta-learning framework using target domain labelled data for meta-testing. They evaluated this approach with multiple domain alignment methods. We used their meta-MME model on SSDA for comparison. LIRR \cite{li2021learning}, optimizes for invariant classifier risk besides invariant representations across the two domains. We compared with two variants, the basic LIRR model and LIRR+CosC, which uses a cosine classifier. CDAC \cite{li2021cross} uses a pseudo-labelling criterion based on ranks of pairwise-similarity between target examples, adversarially maximizing it using the classifier, while minimizing it using the feature representation.

\noindent\textbf{Self Supervision and Domain Adaptation.} 
In the absence of any labelled training data, different self-supervision objectives \cite{carlucci2019domain,  chen2020simple, he2020momentum, gidaris2018unsupervised, misra2020self} have been proposed that can learn semantically meaningful image features for tasks like image classification and object detection.
In domain adaptation some recent approaches \cite{carlucci2019domain, sun2019unsupervised, xu2019self} have used self-supervision tasks as an auxiliary objective to regularize their model. Saito~\etal~\cite{saito2020universal}, used a self-supervised feature space clustering objective for universal domain adaptation. PAC differs from these approaches in that we use rotation prediction to pretrain our feature extractor (see appendix \ref{sec:more-ques} for a comparison with \cite{sun2019unsupervised, xu2019self}). This helps our initial features be more semantically meaningful and better class-wise clustered. In our experiments, we compared this with recent contrastive self-supervision approaches like MoCo \cite{he2020momentum} and SimSiam \cite{chen2021exploring} and found features learnt using rotation prediction to have better properties for our task. This is also in line with the findings of Wallace~\etal~\cite{wallace2020extending}, where rotation prediction was found to be more semantically meaningful compared to other self-supervision objectives, for a range of classification tasks across different datasets.

\vspace{-3mm}

\section{Conclusion}

We showed that consistency regularization and pretraining using rotation prediction are powerful techniques in SSDA. Our method, using simply a combination of these without requiring any domain alignment, could outperform recent state of the art on this task, most of which use adversarial alignment. With our approach we demonstrated that domain alignment is not a necessity for SSDA, and that achieving well-separated target clusters allows for low classifier error with a few labelled examples. We presented a thorough analysis of both the aforementioned techniques showing how they can improve target clustering and why they are better than other options for similar approaches. We hope our analysis can help inform future SSDA work.

\noindent\textbf{Acknowledgements.} This work was supported by the Hariri Institute at Boston University.

\bibliography{submission}

\appendix
\section*{Appendices}

\section{PAC performance with different target shots.} \label{sec:diff-shots}
\begin{figure}[h]
    \centering
    \includegraphics[width=0.75\linewidth,trim=0cm 0cm 0cm 0cm,clip]{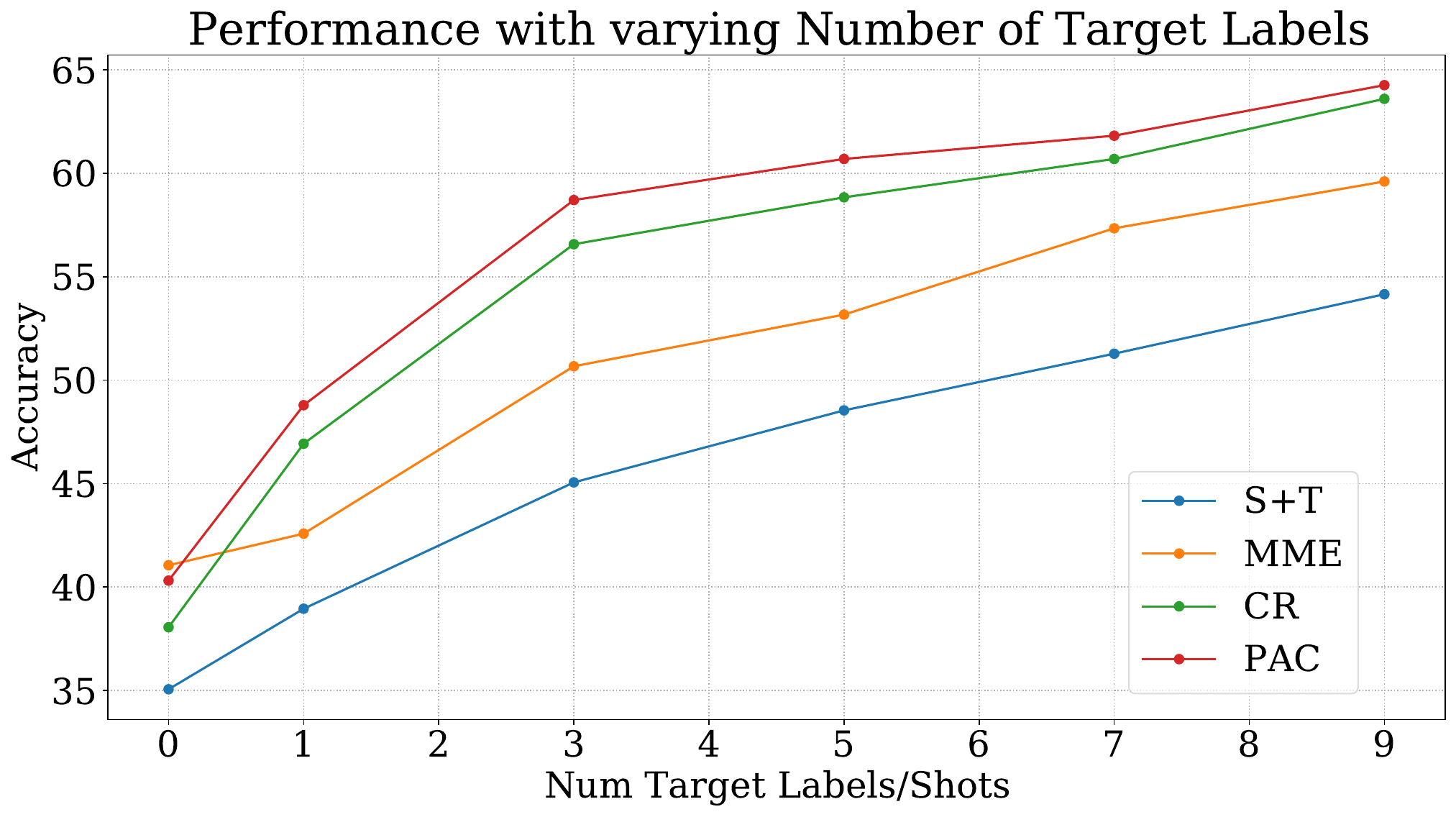}
    \vspace{1mm}
    \caption{Performance with different number of labelled target examples. MME benefiting from domain alignment performs best at 0 shots. With more labelled examples, there is an exponential decrease in target error (Eq. 1), with PAC and CR benefiting most due to better target clustering, \ie, high inter-class divergence $D$}
    \label{fig:diff_shots}
\end{figure}

In Figure \ref{fig:diff_shots}, we plot the target accuracy of 4 methods on the \emph{real} to \emph{clipart} adaptation scenario of Office-Home, for different number of labelled target examples. The method ``CR'' represents the consistency regularization part of PAC, meaning it starts with an Imagenet pretrained backbone, same as S+T and MME \cite{saito2019semi}. We see that with its domain alignment approach, MME performs well at 0 shots. However, along with pretraining using rotation prediction, which has some alignment effect, PAC does not lag far behind. As the number of labelled examples increase, we see all methods enjoy a significant boost in performance, where the error has an exponential relation to the number of labelled examples as indicated by Eq. 1 (main paper). Since PAC and CR have better feature space clustering, \ie, they have a higher inter-class divergence $D$, they see a bigger reduction in error.

\section{More questions} \label{sec:more-ques}

\noindent\textbf{Can consistency regularization fix more errors than MME?} Short answer : yes. In Section 5 of the main paper, we mentioned that consistency regularization, because of the perturbations it makes in image space, can fix errors of the kind that simple conditional entropy minimization, the way it is done in MME, cannot. We validate this hypothesis by training both methods from a randomly initialized feature extractor, where we expect initial features to have a much less meaningful neighborhood in feature space. In Table \ref{tab:from_scratch}, we see a larger gap in the performance of MME starting from a pretrained vs a randomly initialized backbone, which tells us that consistency regularization can fix a lot more errors in the initial feature space than MME. Note that ``Ours (CR)'' method here does not include any rotation pretraining for this comparison.

\begin{table}[h]
\begin{center}
\begin{tabular}{lcc}
\toprule[1.5pt]
Method & Imagenet pt. & Random init. \\
\midrule
MME    & 51.2 & 26.9 \\
Ours (CR)   & 54.1 & 40.0 \\
\bottomrule[1.5pt]
\end{tabular}
\end{center}
\caption{Comparison of MME and our consistency regularization approach on Imagenet pretrained backbone and randomly initialized backbone. Consistency regularization can fix more initial feature space errors than MME.}
\label{tab:from_scratch}
\end{table}

\begin{figure}[h]
    \centering
    \includegraphics[width=0.6\linewidth,trim=0cm 0cm 0cm 0cm,clip]{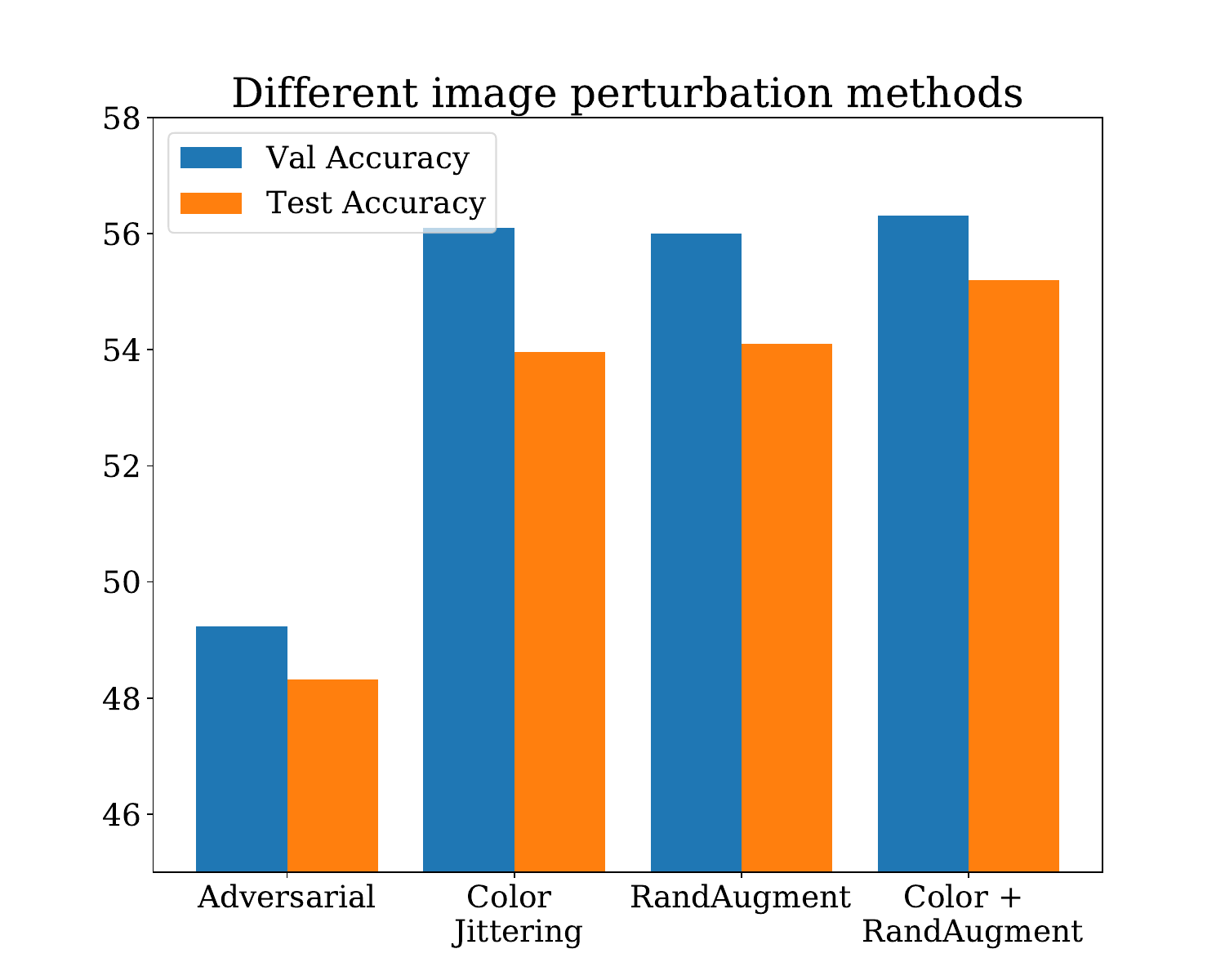}
    \caption{Performance of our method with different augmentation/perturbation methods on real to clipart adaptation of Office-Home. Adversarial perturbation helps, but not as much as image augmentation approaches do. A combination of color jittering and RandAugment performed the best.}
    \label{fig:aug_choice}
\end{figure}

\noindent\textbf{Which perturbation technique is best for consistency?} 
We compared three different image augmentation approaches : RandAugment \cite{cubuk2020randaugment} involves a list of 14 different augmentation schemes like translations, rotations, shears, color/brightness enhancements etc., 2 out of which are chosen randomly anytime an image is augmented. We also evaluated color jittering, since common objects in our datasets are largely invariant to small changes in color. Finally we tried a combination of both, and found that this performed best for our method. Fig \ref{fig:aug_choice} shows the comparison of the final target accuracies achieved using an Alexnet backbone on the \emph{real} to \emph{clipart} adaptation scenario of Office-Home. Besides perturbations based on augmentation, we also evaluated adversarial image perturbation via virtual adversarial training (VAT) \cite{miyato2018virtual}. When using VAT, we found improvements over the simple ``S+T'' method (48.3\% using VAT vs 44.6\% without), but as seen from Fig \ref{fig:aug_choice}, we found this was much lower than image augmentation approaches. This is quite likely because image augmentation imposes a more meaningful neighborhood on images where class labels do not change, while adversarial perturbation does not have this guarantee.

\noindent\textbf{Can pretraining and consistency help other methods?} An indication towards the affirmative is seen when we train MME with pretraining and consistency on the 3-shot \emph{real} to \emph{sketch} scenario of DomainNet using a Resnet-34 backbone. The results are shown in Table \ref{tab:wMME}, where we can see that pretraining and consistency both individually help MME's performance, and their combination helps it the most.

\begin{table}[h]
\begin{center}
\begin{tabular}{ccc}
\toprule[1.2pt]
Rot$^n$ & CR & Accuracy \\
\midrule
& & 61.9 \\
\checkmark & & 65.8 \\
& \checkmark & 70.4 \\
\checkmark & \checkmark & 71.5 \\
\bottomrule[1.2pt]
\end{tabular}
\end{center}
\caption{Pretraining and consistency with MME.}
\label{tab:wMME}
\end{table}

\noindent\textbf{It was explained how pretraining improves initial feature space, but prior work has also used ``pretext'' tasks like rotation prediction alongside classification for training \cite{xu2019self, sun2019unsupervised}. How does pretraining compare to that?} A comparison of this can be found in table \ref{tab:rot_aux}, which reports the 3-shot SSDA target accuracies of the two methods on the DomainNet dataset. As can be seen, pretraining using rotation prediction provides more of a performance benefit as compared to using rotation prediction as an auxiliary task like \cite{xu2019self, sun2019unsupervised}. The latter can help regularize final target classifier training, but likely does not have the benefits that pretraining provides the method via a better initial feature space for training.

\begin{table*}[h]
\begin{center}
\scalebox{0.8}{
\begin{tabular}{l|ccccccc|c}
\toprule[1.2pt]
Method & C2S  & P2C  & P2R  & R2C  & R2P  & R2S  & S2P  & Mean \\
\midrule
S+T + Rot$^n$ pred & 54.7 & 59.5 & 74.1 & 60.4 & 62.3 & 51.8 & 59.2 & 60.3 \\
S+T (Rot$^n$ pred pretrained backbone) & 59.1 & 65.3 & 74.0 & 64.1 & 63.9 & 56.1 & 61.7 & 63.5 \\
\bottomrule[1.2pt]
\end{tabular}
}
\end{center}
\caption{Comparison of rotation prediction for pretraining vs as an auxilliary training task using target accuracies on 3-shot SSDA on different scenarios of DomainNet.}
\label{tab:rot_aux}
\end{table*}

\noindent\textbf{What if pretraining uses rotation prediction only on target?} We train the backbone only on target domain data for pretraining with rotation prediction, and then train it like PAC using consistency regularization. On the 3-shot \emph{real} to \emph{clipart} SSDA scenario of Office-Home using an Alexnet backbone, this achieves a final target accuracy of $57.5$\% compared to $58.9$\% of PAC. This is indicative of target-only rotation prediction helping the initial feature extractor, but not as much as in the case when source domain data is used along with it.  

\begin{table}[h]
\begin{center}
\begin{tabular}{cccc}
\toprule[1.2pt]
Rot$^n$ & CR & \thead{Accuracy \\ (with source)} & \thead{Accuracy \\ (only target)} \\
\midrule
& \checkmark & 56.6 & 35.5 \\
\checkmark & \checkmark & 58.9 & 36.7 \\
\bottomrule[1.2pt]
\end{tabular}
\end{center}
\caption{Ablating source domain information.}
\label{tab:semi-sup}
\end{table}

\noindent\textbf{How big is the role of source domain data in final target performance?} To see this, we train our method with no access to source domain data. This is similar to the semi-supervised learning problem. Target accuracy with only 3 labelled target examples and access to all other unlabelled examples, on the \emph{clipart} domain of Office-Home using an Alexnet backbone, are in the last column of Table \ref{tab:semi-sup}. For reference, the accuracies of our method with source domain data from the \emph{real} domain (\ie R2C adaptation scenario) are provided in the 3$^{rd}$ column.

\section{PAC sensitivity to confidence threshold}
\begin{figure}[t!]
    \centering
    \includegraphics[width=0.95\linewidth,trim=0cm 0cm 0cm 0cm,clip]{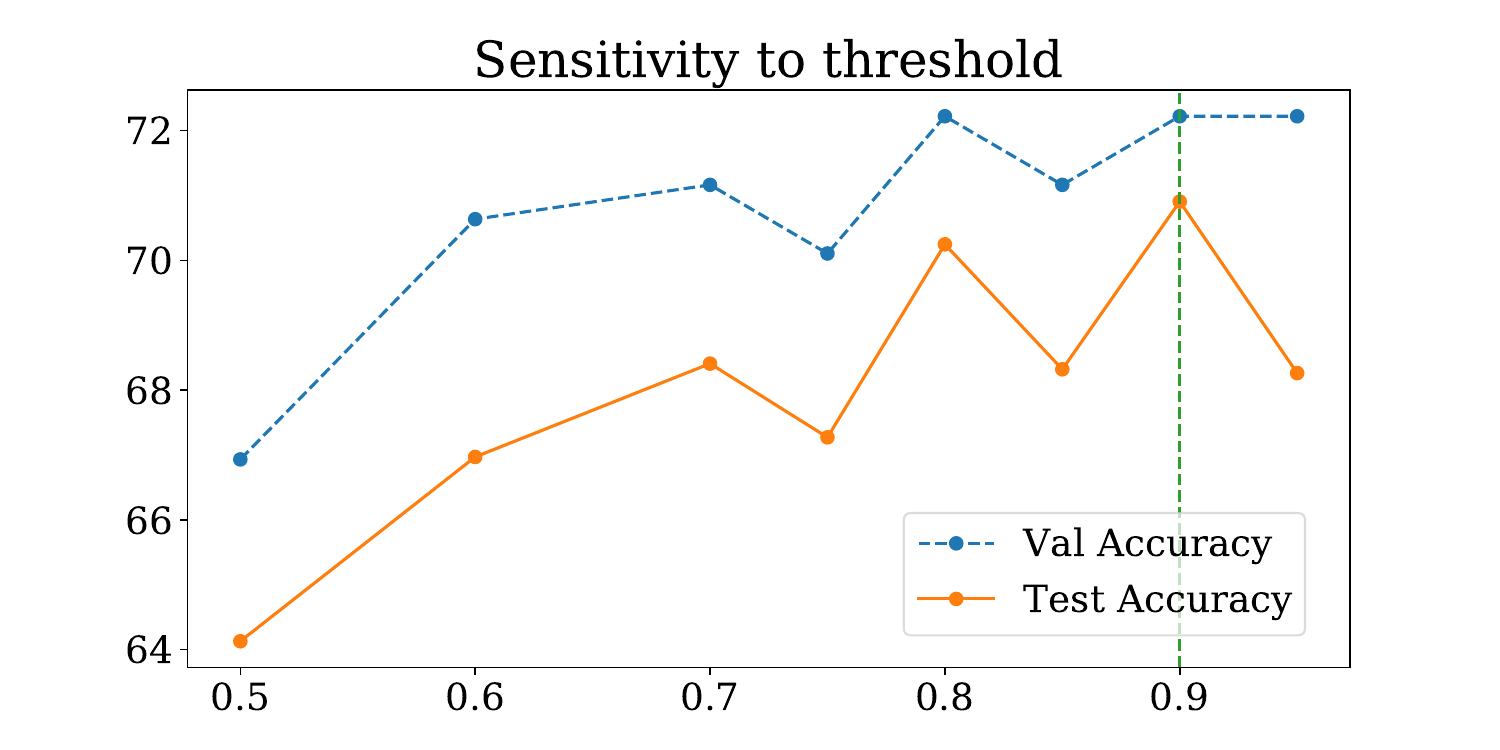}
    \caption{Sensitivity of our method to different thresholds used for consistency regularization. Accuracies reported are on the 3-shot \emph{real} to \emph{sketch} scenario of DomainNet using a Resnet-34 backbone.}
    \label{fig:thres_choice}
\end{figure}

Our consistency regularization approach uses soft targets based on outputs of the classifier only in cases where the confidence of labelling is high. In Fig \ref{fig:thres_choice}, we compare the sensitivity of our method to this threshold. We see that higher confidence thresholds up to 0.9 help final target classification performance. 

\section{Results on Office and Office-Home} \label{sec:full-results}
Office-Home \cite{venkateswara2017Deep} is a dataset with 65 categories of objects found in typical office and home environments. It has 4 different visual domains (Art, Clipart, Product, and Real), and we evaluate our methods on all 12 different adaptation scenarios. The 4 domains have close to 3800 images on average. Office \cite{saenko2010adapting} dataset has objects of 31 different categories in 3 different domains---amazon, webcam and dslr, with approx. 2800, 800 and 500 images respectively. Following \cite{saito2019semi} we evaluated only on the 2 cases with amazon as the target domain, since the other two domains have a lot fewer images. 

Table \ref{tab:office_home_all} shows the results of PAC on the different scenarios of Office-Home, the average accuracy over all these scenarios was also reported in Table 3 in the main paper. Table \ref{tab:office_all} shows the accuracy of PAC on two scenarios of Office. We see that PAC performs comparably to state of the art. It lags behind a little in the 1-shot scenarios as compared to 3-shot ones.

\begin{table}[]
\begin{center}
\scalebox{0.85}{
\begin{tabular}{l|l|cccc}
\toprule[1.5pt]
\multirow{2}{*}{Network}      & \multirow{2}{*}{Method}&\multicolumn{2}{c}{D to A} \ & \multicolumn{2}{c}{W to A}  \\
&&1-shot&3-shot&1-shot&3-shot\\\hline
\multirow{9}{*}{Alexnet} & S+T  & 50.0 & 62.4 & 50.4 & 61.2 \\
                         & DANN & 54.5 & 65.2 & 57.0 & 64.4 \\
                         & ADR  & 50.9 & 61.4 & 50.2 & 61.2 \\
                         & CDAN & 48.5 & 61.4 & 50.2 & 60.3 \\
                         & ENT  & 50.0 & 66.2 & 50.7 & 64.0 \\
                         & MME  & 55.8 & 67.8 & 57.2 & 67.3 \\
                         & APE  & -    & 69.0 & -    & 67.6 \\
                         & BiAT & 54.6 & 68.5 & 57.9 & 68.2 \\
                         & CDAC & \textbf{63.4} & \textbf{70.1} & \textbf{62.8} & \textbf{70.0} \\
                         & PAC & 54.7 & 66.3 & 53.6 & 65.1 \\
\midrule
\multirow{7}{*}{VGG} & S+T  & 68.2 & 73.3 & 69.2 & 73.2 \\
                     & DANN & 70.4 & 74.6 & 69.3 & 75.4 \\
                     & ADR  & 69.2 & 74.1 & 69.7 & 73.3 \\
                     & CDAN & 64.4 & 71.4 & 65.9 & 74.4 \\
                     & ENT  & 72.1 & 75.1 & 69.1 & 75.4 \\
                     & MME  & \textbf{73.6} & \textbf{77.6} & \textbf{73.1} & \textbf{76.3} \\
                     & PAC & 72.4 & 75.6 & 70.2 & 76.0 \\
\bottomrule[1.5pt]
\end{tabular}}
\vspace{2mm}
\caption{Results on Office. We evaluate using the two scenarios where the target domain is \emph{amazon}}
\label{tab:office_all}
\end{center}
\end{table}
\begin{table*}[]
\begin{center}
\scalebox{0.63}{
\begin{tabular}{l|l|cccccccccccc|c}
\toprule[1.5pt] %
Network & Method       &R to C& R to P & R to A & P to R & P to C & P to A & A to P & A to C & A to R & C to R & C to A & C to P & Mean \\\midrule
   \multicolumn{15}{c}{\bf{One-shot}} \\\midrule
\multirow{8}{*}{Alexnet} & S+T  & 37.5 & 63.1 & 44.8 & 54.3 & 31.7 & 31.5 & 48.8 & 31.1 & 53.3 & 48.5 & 33.9 & 50.8 & 44.1 \\
                         & DANN & 42.5 & 64.2 & 45.1 & 56.4 & 36.6 & 32.7 & 43.5 & 34.4 & 51.9 & 51.0 & 33.8 & 49.4 & 45.1 \\
                         & ADR  & 37.8 & 63.5 & 45.4 & 53.5 & 32.5 & 32.2 & 49.5 & 31.8 & 53.4 & 49.7 & 34.2 & 50.4 & 44.5 \\
                         & CDAN & 36.1 & 62.3 & 42.2 & 52.7 & 28.0 & 27.8 & 48.7 & 28.0 & 51.3 & 41.0 & 26.8 & 49.9 & 41.2 \\
                         & ENT  & 26.8 & 65.8 & 45.8 & 56.3 & 23.5 & 21.9 & 47.4 & 22.1 & 53.4 & 30.8 & 18.1 & 53.6 & 38.8 \\
                         & MME  & 42.0 & 69.6 & \textbf{48.3} & \textbf{58.7} & 37.8 & \textbf{34.9} & 52.5 & \textbf{36.4} & \textbf{57.0} & \textbf{54.1} & \textbf{39.5} & \textbf{59.1} & 49.2 \\
                         & BiAT & -    & -    & -    & -    & -    & -    & -    & -    & -    & -    & -    & -    & \textbf{49.6} \\
                         & PAC  & \textbf{49.6} & \textbf{69.8} & 45.9 & 57.5 & \textbf{42.5} & 30.4 & \textbf{53.1} & 35.8 & 51.9 & 48.2 & 26.0 & 57.6 & 47.4 \\
\midrule
\multirow{7}{*}{VGG}     & S+T  & 39.5 & 75.3 & 61.2 & 71.6 & 37.0 & 52.0 & 63.6 & 37.5 & 69.5 & 64.5 & 51.4 & 65.9 & 57.4 \\
                         & DANN & 52.0 & 75.7 & 62.7 & 72.7 & 45.9 & 51.3 & 64.3 & 44.4 & 68.9 & 64.2 & 52.3 & 65.3 & 60.0 \\
                         & ADR  & 39.7 & 76.2 & 60.2 & 71.8 & 37.2 & 51.4 & 63.9 & 39.0 & 68.7 & 64.8 & 50.0 & 65.2 & 57.3 \\
                         & CDAN & 43.3 & 75.7 & 60.9 & 69.6 & 37.4 & 44.5 & 67.7 & 39.8 & 64.8 & 58.7 & 41.6 & 66.2 & 55.9 \\
                         & ENT  & 23.7 & 77.5 & 64.0 & 74.6 & 21.3 & 44.6 & 66.0 & 22.4 & 70.6 & 62.1 & 25.1 & 67.7 & 51.6 \\
                         & MME  & 49.1 & 78.7 & \textbf{65.1} & \textbf{74.4} & 46.2 & \textbf{56.0} & 68.6 & \textbf{45.8} & \textbf{72.2} & \textbf{68.0} & \textbf{57.5} & \textbf{71.3} & \textbf{62.7} \\
                         & PAC  & \textbf{56.4} & \textbf{78.8} & 64.6 & 73.1 & \textbf{54.7} & 55.3 & \textbf{69.8} & 43.5 & 69.5 & 65.3 & 45.3 & 69.6 & 62.2 \\
\midrule
\multicolumn{15}{c}{\bf{Three-shot}} \\\midrule
 \multirow{9}{*}{Alexnet} & S+T  & 44.6 & 66.7 & 47.7 & 57.8 & 44.4 & 36.1 & 57.6 & 38.8 & 57.0 & 54.3 & 37.5 & 57.9 & 50.0 \\
                         & DANN & 47.2 & 66.7 & 46.6 & 58.1 & 44.4 & 36.1 & 57.2 & 39.8 & 56.6 & 54.3 & 38.6 & 57.9 & 50.3 \\
                         & ADR  & 45.0 & 66.2 & 46.9 & 57.3 & 38.9 & 36.3 & 57.5 & 40.0 & 57.8 & 53.4 & 37.3 & 57.7 & 49.5 \\
                         & CDAN & 41.8 & 69.9 & 43.2 & 53.6 & 35.8 & 32.0 & 56.3 & 34.5 & 53.5 & 49.3 & 27.9 & 56.2 & 46.2 \\
                         & ENT  & 44.9 & 70.4 & 47.1 & 60.3 & 41.2 & 34.6 & 60.7 & 37.8 & 60.5 & 58.0 & 31.8 & 63.4 & 50.9 \\
                         & MME  & 51.2 & 73.0 & 50.3 & 61.6 & 47.2 & 40.7 & 63.9 & 43.8 & 61.4 & 59.9 & 44.7 & 64.7 & 55.2 \\
                         & APE  & 51.9 & 74.6 & 51.2 & 61.6 & 47.9 & \textbf{42.1} & \textbf{65.5} & 44.5 & 60.9 & 58.1 & 44.3 & 64.8 & 55.6 \\
                         & BiAT & -    & -    & -    & -    & -    & -    & -    & -    & -    & -    & -    & -    & 56.4 \\
                         & CDAC & 54.9 & \textbf{75.8} & \textbf{51.8} & \textbf{64.3} & 51.3 & \textbf{43.6} & 65.1 & 47.5 & \textbf{63.1} & \textbf{63.0} & \textbf{44.9} & \textbf{65.6} & \textbf{56.8} \\
                         & PAC  & \textbf{58.9} & 72.4 & 47.5 & 61.9 & \textbf{53.2} & 39.6 & 63.8 & \textbf{49.9} & 60.0 & 54.5 & 36.3 & 64.8 & 55.2 \\
\midrule
\multirow{7}{*}{VGG}     & S+T  & 49.6 & 78.6 & 63.6 & 72.7 & 47.2 & 55.9 & 69.4 & 47.5 & 73.4 & 69.7 & 56.2 & 70.4 & 62.9 \\
                         & DANN & 56.1 & 77.9 & 63.7 & 73.6 & 52.4 & 56.3 & 69.5 & 50.0 & 72.3 & 68.7 & 56.4 & 69.8 & 63.9 \\
                         & ADR  & 49.0 & 78.1 & 62.8 & 73.6 & 47.8 & 55.8 & 69.9 & 49.3 & 73.3 & 69.3 & 56.3 & 71.4 & 63.1 \\
                         & CDAN & 50.2 & 80.9 & 62.1 & 70.8 & 45.1 & 50.3 & 74.7 & 46.0 & 71.4 & 65.9 & 52.9 & 71.2 & 61.8 \\
                         & ENT  & 48.3 & 81.6 & 65.5 & 76.6 & 46.8 & 56.9 & 73.0 & 44.8 & \textbf{75.3} & \textbf{72.9} & 59.1 & \textbf{77.0} & 64.8 \\
                         & MME  & 56.9 & \textbf{82.9} & 65.7 & \textbf{76.7} & 53.6 & \textbf{59.2} & 75.7 & 54.9 & \textbf{75.3} & \textbf{72.9} & \textbf{61.1} & 76.3 & 67.6 \\
                         & PAC  & \textbf{63.5} & 82.3 & \textbf{66.8} & 75.8 & \textbf{58.6} & 57.1 & \textbf{75.9} & \textbf{56.7} & 72.2 & 70.5 & 57.7 & 75.3 & \textbf{67.7} \\
\bottomrule[1.5pt]
\end{tabular}}
\end{center}
\caption{Results on all adaptation scenarios of Office-Home.}
\label{tab:office_home_all}

\end{table*}

\section{Experiment details} \label{sec:expt-deets}
All our experiments were implemented in PyTorch \cite{pytorch} using W\&B \cite{wandb} for managing experiments. 

\subsection{PAC experiments}

We used three different backbones for evaluation in different experiments---Alexnet \cite{krizhevsky2012imagenet}, VGG-16 \cite{simonyan2014very} and Resnet-34 \cite{he2016deep}. Our backbones before being trained using the rotation prediction task, are pretrained on the Imagenet \cite{imagenet_cvpr09} dataset, same as other methods used for comparison. While using an Alexnet or VGG-16 feature extractor, we use 1 fully connected layer as the classifier, and while using the Resnet-34 backbone, we use a 2-layer MLP with 512 intermediate nodes. The classifier $C$ uses a temperature parameter set to $0.05$ to sharpen the distribution it outputs using a softmax. For consistency regularization, the confidence threshold $\tau$ was set to 0.9 across all experiments, having validated on the \emph{real} to \emph{sketch} scenario of DomainNet.

Same as \cite{saito2019semi}, we train the models using minibatch-SGD, with $s$ source examples, $s$ labelled target examples and $2s$ unlabelled target examples that the learner ``sees'' at each training step. $s=24$ for the VGG and Resnet backbones, while $s=32$ for Alexnet. The SGD optimizer used a momentum parameter $0.9$ and a weight decay (coefficient of $\ell_2$ regularizer on parameter norm) of $0.0005$.  For all experiments, the parameters of the backbone are updated with a learning rate of $0.001$, while the parameters of the classifier are updated with a learning rate $0.01$. Both of these are decayed as training progresses using a decay schedule similar to  \cite{ganin2015unsupervised}. Learning rate at step $i$ ($\eta_i$) is set as below:

\begin{align*}
    \eta_i = \frac{\eta_0}{\left(1 + 0.0001 \times i \right)^{0.75}}
\end{align*}

For experiments on the Office and Office-Home dataset, we trained PAC using both an Alexnet and a VGG-16 backbone, and the models were trained for 10000 steps with the stopping point chosen using best validation accuracy.

For the experiments on DomainNet, we use both Alexnet and Resnet-34 backbones, while for VisDA-17, we use only Resnet-34. All models in these experiments were trained for 50000 steps, using validation accuracy for determining the best stopping point.

\subsection{Pretraining}
As mentioned above, we pretrain our models for rotation prediction starting from Imagenet pretrained weights. A comparison of PAC with a backbone trained with rotation prediction starting from imagenet pretraining (final target accuracy = $58.9$\%) vs one that does not use any imagenet pretraining (final target accuracy = $43.7$\%), revealed that there is important feature space information in imagenet pretrained weights that rotation prediction could not capture on its own. This comparison was done using an Alexnet on the \emph{real} to \emph{clipart} adaptation scenario of Office-Home.

Following Gidaris~\etal~\cite{gidaris2018unsupervised}, we trained the model on all 4 rotations of a single image in each minibatch. Each minibatch contained $s$ images each from source and target domains, which translates to $4s$ images considering all rotations. 
The Alexnet backbones are trained using a learning rate of $0.01$ and $s = 128$. The Resnet-34 and VGG backbones are both trained using $s = 16$ and a learning rate of $0.001$. We found that beyond a certain point early on in training, the number of steps of training for rotation prediction did not make a big difference to the final task accuracy, and finally the chosen number of training steps was 4000 for Alexnet, 2000 for VGG-16 and 5000 for Resnet-34 backbones.

\subsection{Other Experiments} \label{subsec:expt-deets-other}

\noindent \textbf{MoCo pretraining.} Using the Alexnet backbone, we trained momentum contrast \cite{he2020momentum} for 5000 training steps, where in each step the model saw 32 images each from the \emph{real} and the \emph{clipart} domains of Office-Home. The queue length used for MoCo was 4096 and the momentum parameter was $0.999$. 

\noindent \textbf{SimSiam pretraining.} Using the Alexnet backbone, we trained SimSiam \cite{chen2021exploring} for 200 epochs (or ~6800 training steps) on a mix of the source (\emph{real}) and target (\emph{clipart}) sets of Office-Home, with a batch size of 256.

\noindent \textbf{Virtual Adversarial Training.} For adding a VAT criterion to our model, we closely followed the VAT criterion in VADA \cite{shu2018dirt}. We used a radius of $3.5$ for adversarial perturbations and a coefficient of $0.01$ for the VAT criterion, which is the KL divergence between the outputs of the perturbed and the unperturbed input from the target domain.

\noindent \textbf{Empirical Bhattacharyya Distance Estimate.} We use this estimate to compare target domain inter-class separation in Table 5 of the main paper. For computing an approximation, we made the assumption that features for each class in the target domain are distributed as gaussians with identity covariance and used the closed form Bhattacharyya Distance (BD) between two multivariate gaussians \cite{wiki:bd}. The estimate then reduces to: 
\begin{align*}
    BD = \frac{1}{{K \choose 2}} \sum_{\substack{i, j \in [K] \\ i \neq j}} \frac{1}{8} \norm{\mu_i - \mu_j}_2^2
\end{align*}
where $\mu_i$ is the mean of class $i$ features of images in the target domain ($\D_t \cup \D_u$).

\end{document}